\documentclass[11pt,a4paper,table,xcdraw]{article}
\usepackage[hyperref]{emnlp2020}
\usepackage{times}
\usepackage{latexsym}
\usepackage{amsmath}
\usepackage{amssymb}
\usepackage{graphicx}
\usepackage{hyperref}
\usepackage{booktabs}

\usepackage{xcolor}

\usepackage{multirow}
\usepackage[normalem]{ulem}
\useunder{\uline}{\ul}{}

\usepackage[textsize=tiny]{todonotes}

\newcommand\blfootnote[1]{%
  \begingroup
  \renewcommand\thefootnote{}\footnote{#1}%
  \addtocounter{footnote}{-1}%
  \endgroup
}

\newcommand{\name}[0]{\textsc{tvtropes}}
\newcommand{\imdb}[0]{IMDb}
\newcommand{\goodreads}[0]{Goodreads}
\newcommand{\bvec}[1]{\boldsymbol{#1}}

\usepackage{microtype}

\aclfinalcopy 
\def\aclpaperid{53} 

\title{Analyzing Gender Bias within Narrative Tropes}
\author{\textbf{Dhruvil Gala}$^{\bigstar}$ \hspace{1cm}
\textbf{Mohammad Omar Khursheed}$^{\bigstar}$ \hspace{1cm}
\textbf{Hannah Lerner}$^{}$ \\
\textbf{Brendan O'Connor}$^{}$ \hspace{1cm}
\textbf{Mohit Iyyer}$^{}$
\AND
\normalfont University of Massachusetts Amherst$^{}$\hspace{7mm}\\
\texttt{\{dgala,mkhursheed,hmlerner,brenocon,miyyer\}@umass.edu}\\}

\date{}

\begin{document}
\maketitle
\begin{abstract}
Popular media reflects and reinforces societal biases through the use of \emph{tropes}, which are narrative elements, such as archetypal characters and plot arcs, that occur frequently across media.  In this paper, we specifically investigate \emph{gender bias} within a large collection of tropes. To enable our study, we crawl \url{tvtropes.org}, an online user-created repository that contains 30K tropes associated with 1.9M examples of their occurrences across film, television, and literature. We automatically score the ``genderedness'' of each trope in our \name\ dataset, which enables an analysis of (1) highly-gendered topics within tropes, (2) the relationship between gender bias and popular reception, and (3) how the gender of a work's creator correlates with the types of tropes that they use.
\end{abstract}

\blfootnote{$^{\bigstar}$\textnormal{Authors contributed equally.}}

\section{Introduction}
Tropes are commonly-occurring narrative patterns within popular media. For example, the \href{https://tvtropes.org/pmwiki/pmwiki.php/Main/EvilGenius}{\textbf{evil genius}} trope occurs widely across literature (Lord Voldemort in \emph{Harry Potter}), film (Hannibal Lecter in \emph{The Silence of the Lambs}), and television (Tywin Lannister in \emph{Game of Thrones}). Unfortunately, many tropes exhibit gender bias\footnote{Our work explores gender bias across two identities: cisgender male and female. The lack of reliable lexicons limits our ability to explore bias across other gender identities, which should be a priority for future work.}, either explicitly through stereotypical generalizations in their definitions, or implicitly through biased representation in their usage that exhibits such stereotypes. Movies, TV shows, and books with stereotypically gendered tropes and biased representation reify and reinforce gender stereotypes in society~\citep{rowe2011unruly,gupta2008mis,leonard2006not}. While \textbf{evil genius} is not an explicitly gendered trope (as opposed to, for example, \href{https://tvtropes.org/pmwiki/pmwiki.php/Main/WomenAreWiser}{\textbf{women are wiser}}), the online \url{tvtropes.org} 
repository contains 108 male and only 15 female instances of \textbf{evil genius} across film, TV, and literature.

To quantitatively analyze gender bias within tropes, we collect \name, a large-scale dataset that contains 1.9M examples of 30K tropes in various forms of media. 
We augment our dataset with metadata from \href{https://developer.imdb.com}{\imdb} (year created, genre, rating of the film/show) and \href{http://www.goodreads.com}{\goodreads} (author, characters, gender of the author), which enable the exploration of how trope usage differs across contexts.   

Using our dataset, we develop a simple method based on counting pronouns and gendered terms to compute a \emph{genderedness score} for each trope. Our computational analysis of tropes and their genderedness reveals the following:

\begin{itemize}
    \setlength\itemsep{0em}
    \item \textbf{Genre impacts genderedness:} Media related to sports, war, and science fiction rely heavily on male-dominated tropes, while romance, horror, and musicals lean female. 
    \item \textbf{Male-leaning tropes exhibit more topical diversity:} Using LDA, we show that male-leaning tropes exhibit higher topic diversity (e.g., science, religion, money) than female tropes, which contain fewer distinct topics (often related to sexuality and maternalism).
    \item \textbf{Low-rated movies contain more gendered tropes:} Examining the most informative features of a classifier trained to predict \imdb\ ratings for a given movie reveals that gendered tropes are strong predictors of low ratings.
    \item \textbf{Female authors use more diverse gendered tropes than male authors:} Using author gender metadata from Goodreads, we show that female authors incorporate a more diverse set of female-leaning tropes into their works. 
\end{itemize}

\noindent
Our dataset and experiments complement existing social science literature that qualitatively explore gender bias in media~\citep{lauzen2019sa}. We publicly release \name\footnote{\url{http:/github.com/dhruvilgala/tvtropes}} to facilitate future research that computationally analyzes bias in media. 

\section{Collecting the \name\ dataset}
We crawl \url{TVTropes.org} to collect a large-scale dataset of 30K tropes and 1.9M examples of their occurrences across 40K works of film, television, and literature. We then connect our data to metadata from \imdb\ and \goodreads\ to augment our dataset and enable analysis of gender bias.
            
\subsection{Collecting a dataset of tropes}
Each trope on the website contains a \emph{description} as well as a set of \emph{examples} of the trope in different forms of media. Descriptions normally consist of multiple paragraphs (277 tokens on average), while examples are shorter (63 tokens on average). We only consider titles from film, TV, and literature, excluding other forms of media, such as web comics and video games.
We focus on the former because we can pair many titles with their \imdb\ and \goodreads\ metadata.
Table~\ref{tab:stats} contains statistics of the \name\ dataset.



\begin{table}[t!]
\centering
\resizebox{\columnwidth}{!}{
\begin{tabular}{@{}lrcrrr@{}}
\toprule
                     & \multicolumn{2}{l}{\textbf{Titles (w/ metadata)}} & \multicolumn{1}{l}{\textbf{Tropes}} & \multicolumn{1}{l}{\textbf{Examples}}  \\ \midrule
\textbf{Literature}  & 15,495 & (5,208)                               & 27,229                               & 679,618 \\
\textbf{Film}        & 17,019 &(8,816)                               & 27,450                               & 751,594 \\
\textbf{TV}  & 7,921 &(4,192)                             & 27,134                               & 488,632 \\
\textbf{Total}       & 40,435 &(18,216)                               & 29,457                               & 1,919,844 \\                                    
\bottomrule
\end{tabular}}
\caption {Statistics of \name.}
\label{tab:stats}
\end{table}

\subsection{Augmenting \name\ with metadata}
We attempt to match\footnote{We match by both the work's title and its year of release to avoid duplicates.} each film and television listed in our dataset with publicly-available \imdb\ metadata, which includes year of release, genre, director and crew members, and average rating. Similarly, we match our literature examples with metadata scraped from \goodreads, which includes author names, character lists, and book summaries. We additionally manually annotate author gender from \goodreads\ author pages. The second column of Table~\ref{tab:stats} shows how many titles were successfully matched with metadata through this process.

\subsection{Who contributes to \name?}
One limitation of any analysis of social bias on \name\ is that the website may not be representative of the true distribution of tropes within media. There is a confounding \textit{selection bias}---the media in \name\ is selected by the users who maintain the \url{tvtropes.org} resource. To better understand the demographics of contributing users,
we scrape the pages of the 15K contributors, many of which contain unstructured biography sections. We search for biographies that contain tokens related to gender and age, and then we manually extract the reported gender and age for a sample of 256 contributors.\footnote{We note that some  demographics may be more inclined to report age and gender information than others.} The median age of these contributors is 20, while 64\% of them are male, 33\% female and 3\% bi-gender, genderfluid, non-binary, trans, or agender. We leave exploration of whether user-reported gender correlates with properties of contributed tropes to future work.

\section{Measuring trope genderedness}
We limit our analysis to male and female genders, though we are keenly interested in examining the correlations of other genders with trope use.
We devise a simple score for trope genderedness that relies on matching tokens to male and female lexicons\footnote{The gender-balanced lexicon is obtained from \citet{gnglove} and comprises 222 male-female word pairs.} used in prior work~\cite{DBLP:journals/corr/BolukbasiCZSK16a, gnglove} and include gendered pronouns, possessives (\emph{his}, \emph{her}), occupations (\emph{actor}, \emph{actress}), and other gendered terms.
We validate the effectiveness of the lexicon in capturing genderedness by annotating 150 random examples of trope occurrences as male (86), female (23), or N/A (41). N/A represents examples that do not capture any aspect of gender. We then use the lexicon to classify each example as male (precision $= 0.85$, recall $= 0.86$, and F1 score $= 0.86$) or female (precision $= 0.72$, recall $= 0.78$, and F1 score $= 0.75$).

To measure genderedness, for each trope $i$, we concatenate the trope's description with all of the trope's examples to form a document $X_i$.
Next, we tokenize, preprocess, and lemmatize $X_i$ using NLTK \cite{nltk}. We then compute the number of tokens in $X_i$ that match the male lexicon, $m(X_i)$, and  the female lexicon, $f(X_i)$. 
We also compute $m(\name)$ and $f(\name)$, the total number of matches for each gender across all trope documents in the corpus. The \emph{raw genderedness score} of trope $i$ is the ratio $d_i=$
\begin{equation*}\begin{small}
    \underbrace{\frac{f(X_i)}{f(X_i) + m(X_i)}}_{r_i} 
\Bigg/ 
\underbrace{\frac{f(\name)}{f(\name) + m(\name)}}_{r_{\name}}.
\end{small}
\end{equation*}


\noindent 
This score is a trope's proportion of female tokens among gendered tokens ($r_i$), normalized by the global ratio in the corpus ($r_{\name}$=0.32).
If $d_i$ is high, trope $i$ contains a larger-than-usual proportion of female words. 

We finally calculate the the \emph{genderedness score} $g_i$ as $d_i$'s normalized $z$-score.\footnote{$g_i \approx 0$ when $r_i=r_{\name}$} This results in scores from $-1.84$ (male-dominated) to $4.02$ (female-dominated). For our analyses, we consider tropes with genderedness scores outside of $[-1, 1]$ (one standard deviation) to be highly gendered (see Table~\ref{tab:my-table} for examples).

While similar to methods used in prior work~\cite{DBLP:journals/corr/GarciaWG14}, our genderedness score is limited by its lexicon and susceptible to gender generalization and explicit marking \cite{hitti-etal-2019-proposed}. We leave exploration of more nuanced methods of capturing trope genderedness~\citep{ananya-etal-2019-genderquant} to future work.

\begin{table}[t!]
\centering
\resizebox{!}{1.4cm}{%
\begin{tabular}{lr lr}\toprule
\textbf{Male Tropes} & $g$ & \textbf{Female Tropes} &  $g$ \\ \midrule
\href{https://tvtropes.org/pmwiki/pmwiki.php/Main/MotivatedByFear}{\textbf{Motivated by Fear}}    & -1.8       & \href{https://tvtropes.org/pmwiki/pmwiki.php/Main/MsFanservice}{\textbf{Ms. Fanservice}}          & 3.4        \\
\href{https://tvtropes.org/pmwiki/pmwiki.php/Main/RobotWar}{\textbf{Robot War}}            & -1.6       & \href{https://tvtropes.org/pmwiki/pmwiki.php/Main/Socialite}{\textbf{Socialite}}              & 3.1        \\
\href{https://tvtropes.org/pmwiki/pmwiki.php/Main/CureForCancer}{\textbf{Cure for Cancer}}      & -1.5       & \href{https://tvtropes.org/pmwiki/pmwiki.php/Main/DamselInDistress}{\textbf{Damsel in Distress}}     & 2.7        \\
\href{https://tvtropes.org/pmwiki/pmwiki.php/Main/EvilGenius}{\textbf{Evil Genius}}          & -1.3       & \href{https://tvtropes.org/pmwiki/pmwiki.php/Main/HotScientist}{\textbf{Hot Scientist}}          & 2.2        \\
\href{https://tvtropes.org/pmwiki/pmwiki.php/Main/GrandFinale}{\textbf{Grand Finale}}         & -1.2       & \href{https://tvtropes.org/pmwiki/pmwiki.php/Main/DitzySecretary}{\textbf{Ditzy Secretary}}         & 2.0       \\ \bottomrule
\end{tabular}%
}
\caption{Instances of highly-gendered tropes.}
\label{tab:my-table}
\end{table}

\section{Analyzing gender bias in \name}
Having collected \name\ and linked each trope with metadata and genderedness scores, we now turn to characterizing how gender bias manifests itself in the data. We explore (1) the effects of genre on genderedness, (2) what kinds of topics are used in highly-gendered tropes, (3) what tropes contribute most to \imdb\ ratings, and (4) what types of tropes are used more commonly by authors of one gender than another. 

                        
\subsection{Genderedness across genre}
\begin{figure}[t!]
    \centering
    \includegraphics[width=\columnwidth]{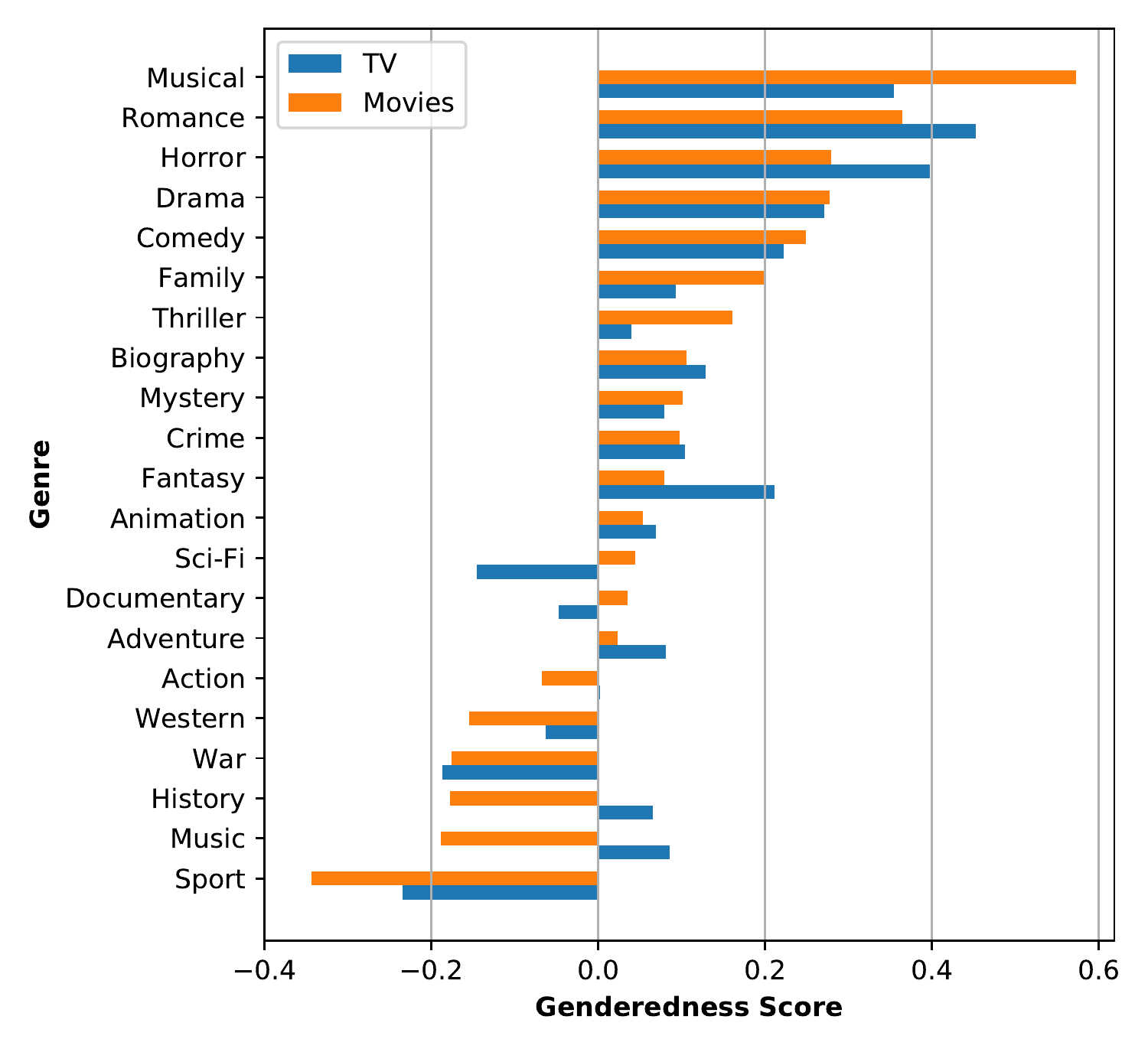}
    \caption{Genderedness across film and TV genres.}
    \label{fig:genderedness_film_tv}
\end{figure}

We can examine how genderedness varies by genre. Given the set of all movies and TV shows in \name\ that belong to a particular genre, we extract the set of all tropes used in these works. Next, we compute the average genderedness score of all of these tropes. Figure~\ref{fig:genderedness_film_tv}
shows that media about sports, war, and science fiction contain more male-dominated tropes, while musicals, horror, and romance shows are heavily oriented towards female tropes, which is corroborated by social science literature~\citep{lauzen2019sa}.


\subsection{Topics in highly-gendered tropes}
To find common topics in highly-gendered male or female tropes, we run latent Dirichlet analysis~\citep{blei2003latent} on a subset of highly-gendered trope descriptions and examples\footnote{We use Gensim's LDA library~\citep{rehurek_lrec}.} with 75 topics. We filter out tropes whose combined descriptions and examples (i.e., $X_i$) have fewer than 1K tokens, and then we further limit our training data to a balanced subset of the 3,000 most male and female-leaning tropes using our genderedness score. After training, we compute a gender ratio for every \emph{topic}: given a topic $t$, we identify the set of all tropes for which $t$ is the most probable topic, and then we compute the ratio of female-leaning to male-leaning tropes within this set. 

We observe that 45 of the topics are skewed towards male tropes, while 30 of them favor female tropes, suggesting that male-leaning tropes cover a larger, more diverse set of topics than female-leaning tropes. Table~\ref{tab:topics} contains specific examples of the most gendered male and female topics. This experiment, in addition to a qualitative inspection of the topics, reveals that female topics (maternalism, appearance, and sexuality) are less diverse than male topics (science, religion, war, and money). Topics in highly-gendered tropes capture all three dimensions of sexism proposed by \newcite{glick} -- female topics about motherhood and pregnancy display gender differentiation, topics about appearance and nudity can be attributed to heterosexuality, while male topics about money and strength capture paternalism. The bias captured by these topics, although unsurprising given previous work \cite{DBLP:journals/corr/BolukbasiCZSK16a}, serves as a sanity check for our metric and provides further evidence of the limited diversity in female roles~\citep{lauzen2019sa}.
\begin{table}[t!]
\centering
\resizebox{\columnwidth}{!}{\begin{tabular}{@{}cc@{}}
\toprule
                                 & \textbf{Topics (Most salient terms)}                                                             \\ \midrule
\multirow{5}{*}{\textbf{Male}}   & ship earth planet technology system build weapon destroy alien      \\ \cmidrule(l){2-2} 
                                 & super strong strength survive slow damage speed hulk punch armor            \\ \cmidrule(l){2-2} 
                                 & god jesus religion church worship bible believe angel heaven belief         \\ \cmidrule(l){2-2} 
                                 & skill rule smart training problem student ability level teach genius        \\ \cmidrule(l){2-2} 
                                 & money rich gold steal company city sell business criminal wealthy           \\ \midrule
\multirow{5}{*}{\textbf{Female}} & relationship married marry wife marriage together husband wedding    \\ \cmidrule(l){2-2} 
                                 & beautiful blonde attractive beauty describe tall brunette eyes ugly \\ \cmidrule(l){2-2} 
                                 & naked sexy fanservice shower nude cover strip pool bikini shirt             \\ \cmidrule(l){2-2} 
                                 & parent baby daughter pregnant birth kid die pregnancy raise adult           \\ \cmidrule(l){2-2} 
                                 & food drink eating cook taste weight drinking chocolate wine         \\ \bottomrule 
\end{tabular}}
\caption{Topic assignments in highly-gendered tropes.}
\label{tab:topics}
\end{table}

\subsection{Identifying implicitly gendered tropes}
We identify \emph{implicitly gendered tropes}~\cite{glick}---tropes that are not defined by gender but nevertheless have high genderedness scores---by identifying a subset of 3500 highly-gendered tropes whose titles do not contain gendered tokens.\footnote{This process contains noise due to our small lexicon: a few explicitly gendered, often problematic tropes such as \href{https://tvtropes.org/pmwiki/pmwiki.php/Main/AbsoluteCleavage}{\textbf{absolute cleavage}} are not filtered out.}  A qualitative analysis reveals that tropes containing the word ``genius'' (\href{https://tvtropes.org/pmwiki/pmwiki.php/Main/ImpossibleGenius}{\textbf{impossible genius}}, \href{https://tvtropes.org/pmwiki/pmwiki.php/Main/GibberingGenius}{\textbf{gibbering genius}}, \href{https://tvtropes.org/pmwiki/pmwiki.php/Main/EvilGenius}{\textbf{evil genius}}) and ``boss'' (\href{https://tvtropes.org/pmwiki/pmwiki.php/Main/BeleagueredBoss}{\textbf{beleaguered boss}}, \href{https://tvtropes.org/pmwiki/pmwiki.php/Main/StupidBoss}{\textbf{stupid boss}}) lean heavily male. There are interesting gender divergences within a high-level topic: within ``evil'' tropes, male-leaning tropes are diverse (\href{https://tvtropes.org/pmwiki/pmwiki.php/Main/NecessarilyEvil}{\textbf{necessarily evil}}, \href{https://tvtropes.org/pmwiki/pmwiki.php/Main/EvilCorporation}{\textbf{evil corporation}}, \href{https://tvtropes.org/pmwiki/pmwiki.php/Main/EvilArmy}{\textbf{evil army}}), while female tropes focus on sex  (\href{https://tvtropes.org/pmwiki/pmwiki.php/Main/SexIsEvil}{\textbf{sex is evil}}, \href{https://tvtropes.org/pmwiki/pmwiki.php/Main/EvilEyeshadow}{\textbf{evil eyeshadow}}, \href{https://tvtropes.org/pmwiki/pmwiki.php/Main/EvilIsSexy}{\textbf{evil is sexy}}). 
\subsection{Using tropes to predict ratings}
Are gendered tropes predictive of media popularity? We consider three roughly equal-sized bins of IMDb ratings (Low, Medium, and High).\footnote{Low: (0-6.7], Medium: (6.7-7.7], High: (7.7-10]} For each IMDb-matched title in \name, we construct a binary vector $\bvec{z} \in \{0,1\}^T$, where $T$ is the number of unique tropes in our dataset.\footnote{We consider titles and tropes with 10+ examples.} We set $\bvec{z}_i$ to 1 if trope $i$ occurs in the movie, and 0 otherwise. Tropes are predictive of ratings: a logistic regression classifier\footnote{\label{note1}We implement the classifier in scikit-learn~\citep{scikit-learn} with L-BFGS solver, L2 regularization, inverse regularization strength C=1.0 and an 80-20 train-test split.} achieves 55\% test accuracy with this method, well over the majority class baseline of 36\%.
Table~\ref{tab:trope_pred_rating} contains the most predictive gendered tropes for each class; interestingly, low-rated titles have a much higher average \emph{absolute} genderedness score (0.73) than high-rated ones (0.49), providing interesting contrast to the opposing conclusions drawn by \citet{boyle2014gender}. While IMDB ratings offer a great place to start in correlating public perception with genderedness in tropes, we may be double-dipping into the same pool of internet movie reviewers as \name. We leave further exploration of correlating gendered tropes with box office results, budgets, awards, etc. for future work.

\begin{table}[t!]
\centering
\resizebox{!}{1.355cm}{%
\begin{tabular}{
>{\columncolor[HTML]{FFFFFF}}c 
>{\columncolor[HTML]{FFFFFF}}c 
>{\columncolor[HTML]{FFFFFF}}c 
>{\columncolor[HTML]{FFFFFF}}c }
\toprule
\multicolumn{2}{c}{\cellcolor[HTML]{FFFFFF}{\color[HTML]{000000} \textbf{High Rated}}} &
  \multicolumn{2}{c}{\cellcolor[HTML]{FFFFFF}{\color[HTML]{000000} \textbf{Low Rated}}} \\ \midrule
{\color[HTML]{000000} \textbf{Trope}} &
  {\color[HTML]{000000} \textbf{$g$}} &
  {\color[HTML]{000000} \textbf{Trope}} &
  {\color[HTML]{000000} \textbf{$g$}} \\ \midrule
{\color[HTML]{000000} \href{https://tvtropes.org/pmwiki/pmwiki.php/Main/EdutainmentShow}{\textbf{Edutainment Show}}}        & {\color[HTML]{000000} 1.2}  & 
{\color[HTML]{000000} \href{https://tvtropes.org/pmwiki/pmwiki.php/Main/AlphaBitch}{\textbf{Alpha Bitch}}} & {\color[HTML]{000000} 2.7} \\
{\color[HTML]{000000} 
\href{https://tvtropes.org/pmwiki/pmwiki.php/Main/CookingStories}{\textbf{Cooking Stories}}} & {\color[HTML]{000000} 1.0} &
 {\color[HTML]{000000}
  \href{https://tvtropes.org/pmwiki/pmwiki.php/Main/SexyBacklessOutfit}{\textbf{Sexy Backless Outfit}}} &
  {\color[HTML]{000000} 2.6}\\
  {\color[HTML]{000000} \href{https://tvtropes.org/pmwiki/pmwiki.php/Main/BritishBrevity}{\textbf{British Brevity}}}             & {\color[HTML]{000000} -0.9} & {\color[HTML]{000000} \href{https://tvtropes.org/pmwiki/pmwiki.php/Main/Fanservice}{\textbf{Fanservice}}}      & {\color[HTML]{000000} 1.9} \\
{\color[HTML]{000000} \href{https://tvtropes.org/pmwiki/pmwiki.php/Main/WeddingSmashers}{\textbf{Wedding Smashers}}} & {\color[HTML]{000000} 0.8} &
  {\color[HTML]{000000} \href{https://tvtropes.org/pmwiki/pmwiki.php/Main/ShowerScene}{\textbf{Shower Scene}}} &
  {\color[HTML]{000000} 1.4} \\
{\color[HTML]{000000} \href{https://tvtropes.org/pmwiki/pmwiki.php/Main/JustFollowingOrders}{\textbf{Just Following Orders}}} &
  {\color[HTML]{000000} -0.7} & {\color[HTML]{000000} \href{https://tvtropes.org/pmwiki/pmwiki.php/Main/SwordAndSandal}{\textbf{Sword and Sandal}}}      & {\color[HTML]{000000} -1.0}  
\\ \bottomrule
\end{tabular}%
}\caption{Gendered tropes predictive of \imdb\ rating.}
\label{tab:trope_pred_rating}
\end{table}

\subsection{Predicting author gender from tropes}
We predict the author gender\footnote{We annotate the author gender label by hand, to prevent misgendering based on automated detection methods, and we would also like to further this research by expanding our Goodsreads scrape to include non-binary authors.} by training a classifier for 2521 \goodreads\ authors based on a binary feature vector encoding the presence or absence of tropes in their books. We achieve an accuracy of 71\% on our test set (majority baseline is 64\%). Interestingly, the top 50 tropes most predictive of male authors have an average genderedness of 0.04, while those most correlated with female authors have an average of 0.89, indicating that books by female authors contain more female-leaning tropes. Eighteen female-leaning tropes ($g_i > 1$), varying in scope from the non-traditional \href{https://tvtropes.org/pmwiki/pmwiki.php/Main/FeministFantasy}{\textbf{feminist fantasy}} to the more stereotypical \href{https://tvtropes.org/pmwiki/pmwiki.php/Main/HairOfGoldHeartOfGold}{\textbf{hair of gold heart of gold}}, are predictive of female authors. In contrast, only two such character-driven female-dominated tropes are predictive of male authors; the stereotypical \href{https://tvtropes.org/pmwiki/pmwiki.php/Main/UndressingTheUnconscious}{\textbf{undressing the unconscious}} and \href{https://tvtropes.org/pmwiki/pmwiki.php/Main/FirstGirlWins}{\textbf{first girl wins}}; see Table~\ref{tab:trope_pred_gender} for more. Furthermore, out of 115K examples of tropes in female-authored books, 17K are highly female, while just 2.2K are male-dominated. Since many of these gendered tropes are character-driven, this implies wider female representation in such gendered instances, previously shown in Scottish crime fiction~\citep{scottish-change-female-authors}. Overall, female authors frequently use both stereotypical and non-stereotypical female-oriented tropes, while male authors limit themselves to more stereotypical kinds. However, it is important to note the double selection bias at play in both selecting which books are actually published, as well as which published books are reviewed on \goodreads. While there are valid ethical concerns with a task that attempts to predict gender, this task only analyzes the tropes most predictive of author gender, and the classifier is not used to do inference on unlabelled data or as a way to identify an individual's gender. 

\begin{table}[t!]
\centering
\resizebox{\columnwidth}{!}{%
\begin{tabular}{
>{\columncolor[HTML]{FFFFFF}}c 
>{\columncolor[HTML]{FFFFFF}}c 
>{\columncolor[HTML]{FFFFFF}}c 
>{\columncolor[HTML]{FFFFFF}}c }
\toprule
\multicolumn{2}{c}{\cellcolor[HTML]{FFFFFF}{\color[HTML]{000000} \textbf{Male Author}}} &
  \multicolumn{2}{c}{\cellcolor[HTML]{FFFFFF}{\color[HTML]{000000} \textbf{Female Author}}} \\  \midrule
\multicolumn{1}{c}{\cellcolor[HTML]{FFFFFF}{\color[HTML]{000000} \textbf{Trope}}} &
  \multicolumn{1}{c}{\cellcolor[HTML]{FFFFFF}{\color[HTML]{000000} \textbf{$g$}}} &
  \multicolumn{1}{c}{\cellcolor[HTML]{FFFFFF}{\color[HTML]{000000} \textbf{Trope}}} &
  \multicolumn{1}{c}{\cellcolor[HTML]{FFFFFF}{\color[HTML]{000000} \textbf{$g$}}} \\ \midrule
{\color[HTML]{000000} \href{https://tvtropes.org/pmwiki/pmwiki.php/Main/UndressingTheUnconscious}{\textbf{Undressing the Unconscious}}} &
  {\color[HTML]{000000} 1.3}  &
  {\color[HTML]{000000} \href{https://tvtropes.org/pmwiki/pmwiki.php/Main/CoolOldLady}{\textbf{Cool Old Lady}}} &
  {\color[HTML]{000000} 2.7} \\
  {\color[HTML]{000000} \href{https://tvtropes.org/pmwiki/pmwiki.php/Main/FirstGirlWins}{\textbf{First Girl Wins}}} &
  {\color[HTML]{000000} 1.3} &
  {\color[HTML]{000000} 
  \href{https://tvtropes.org/pmwiki/pmwiki.php/Main/PluckyGirl}{\textbf{Plucky Girl}}} &
  {\color[HTML]{000000} 2.5} \\
  {\color[HTML]{000000} \href{https://tvtropes.org/pmwiki/pmwiki.php/Main/DidNotGetTheGirl}{\textbf{Did Not Get the Girl}}} &
  {\color[HTML]{000000} 0.9} &
  {\color[HTML]{000000} \href{https://tvtropes.org/pmwiki/pmwiki.php/Main/FeministFantasy}{\textbf{Feminist Fantasy}}} &
  {\color[HTML]{000000} 2.2} \\
{\color[HTML]{000000} \href{https://tvtropes.org/pmwiki/pmwiki.php/Main/GodIsEvil}{\textbf{God is Evil}}} &
  {\color[HTML]{000000} -0.8} &
  {\color[HTML]{000000} \href{https://tvtropes.org/pmwiki/pmwiki.php/Main/YoungAdultLiterature}{\textbf{Young Adult Literature}}} &
  {\color[HTML]{000000} 1.7} \\

  {\color[HTML]{000000} \href{https://tvtropes.org/pmwiki/pmwiki.php/Main/RetiredBadass}{\textbf{Retired Badass}}} &
  {\color[HTML]{000000} -0.8}&
  {\color[HTML]{000000} \href{https://tvtropes.org/pmwiki/pmwiki.php/Main/ExtremelyProtectiveChild}{\textbf{Extremely Protective Child}}} &
  {\color[HTML]{000000} 1.2} \\ \bottomrule
\end{tabular}%
}\caption{Gendered tropes predictive of author gender.}
\label{tab:trope_pred_gender}
\end{table}


\section{Related Work}
Our work builds on computational research analyzing gender bias. Methods to measure gender bias include using contextual cues to develop probabilistic estimates~\citep{ananya-etal-2019-genderquant}, and using gender directions in word embedding spaces~\citep{DBLP:journals/corr/BolukbasiCZSK16a}. Other work engages directly with \url{tvtropes.org}:
\newcite{dbtropes} build a wrapper for the website, but perform no analysis of its content.~\citet{garcaortega} create PicTropes: a limited dataset of 5,925 films from the website.~\citet{bamman-etal-2013-learning} collect a set of 72 character-based tropes, which they then use to evaluate induced character personas, and~\citet{lee-etal-2019-understanding} use data crawled from the website to explore different sexism dimensions within TV and film. 

Analyzing bias through tropes is a popular area of research within social science. \citet{hansen2018can} focus in on the titular princess character in the video game The Legend of Zelda as an example of the Damsel in Distress trope. \citet{doi:10.1080/10646175.2011.546738} study the  development and representation in popular culture of the Casino Indian and Ignoble Savage tropes.

The usage of biased tropes is often attributed to the lack of equal representation both on and off the screen. The \emph{Geena Davis Inclusion Quotient}~\citep{google}  quantifies the speaking time and importance of characters in films, and finds that male characters have nearly twice the presence of female characters in award-winning films. In contrast, our analysis looks specifically at tropes, which may not correlate directly with speaking time. \citet{lauzen2019sa} provides valuable insight into representation among film writers, directors, crew members, etc. \citet{doi:10.1080/14680777.2019.1667059} study an ongoing increase in the representation of women in independent productions on television, many of which focus on feminist content. 

\section{Future Work}
We believe that the \name\ dataset can be used to further research in a variety of areas. We envision setting up a task involving trope detection from raw movie scripts or books; the resulting classifier, beyond being useful for analysis, could also be used by media creators to foster better representation during the writing process. There is also the possibility of using the large number of examples we collect in order to generate augmented training data or adversarial data for tasks such as coreference resolution  in a gendered context~\citep{rudinger2018gender}. The expansion of our genderedness metric to include non-binary gender idenities, which in turn would involve creating similar lexicons as we use, is an important area for further exploration. 

It would also be useful to gain further understanding of the multiple online communities that contribute information about popular culture; for example, an analysis of possible overlap in contributors to \name\ and IMDb could better account for sampling bias when analyzing these datasets.

\section{Acknowledgements}
We would like to thank Jesse Thomason for his valuable advice. 

\bibliography{acl2020}
\bibliographystyle{acl_natbib}

\end{document}